\documentclass[letterpaper, 10 pt, conference]{ieeeconf}  

\IEEEoverridecommandlockouts                              

\usepackage{graphics} 
\usepackage{epsfig} 
\usepackage{float}

\title{\LARGE \bf
A Reconfigurable Hybrid Actuator with Rigid and Soft Components
}

\author{Yaohui Chen$^{}$, Sing Le$^{}$, Qiao Chu Tan$^{}$, Oscar Lau$^{}$, Fang Wan$^{}$, Chaoyang Song$^{*}$, \textit{Member, IEEE}
\thanks{*Correspondence author}
\thanks{Y. Chen, L. Sing, Q. Tan, and O. Lau are students with the Faculty of Engineering, Monash University, Clayton, VIC, 3168, Australia (email: {\small yaohui.chen@monash.edu}; {\small \{sle11, qctan1, olau2\}@student.monash.edu}.)
\newline \indent F. Wan was with Nanyang Technological University, Singapore and now is an independent researcher. (email:  {\small sophie.fwan@gmail.com})
\newline \indent C. Song is Lecturer with the Faculty of Engineering, Monash University, Clayton, VIC, 3168, Australia (phone: +61-03-99052994, email: 
{\small songcy@ieee.org}).}
}
\begin{document}

\maketitle
\thispagestyle{empty}
\pagestyle{empty}

\begin{abstract}
Classical rigid-bodied robotic systems are presented with proven success in theoretical development and industrial applications, are recently challenged by the emergence of soft robotics due to a growing need in physical human-robot interactions (pHRI), such as wearable devices, medical robots, personal robots, etc. In this paper, we present the design and fabrication of a robust, hybrid bending actuator build from both rigid and soft components inspired by crustaceans, where its bending radius and axis can be mechanically programmed through the selective activation of the rigid exterior joints, actuated by the soft actuators inside. The hybrid actuator was experimentally measured in terms of bending and force tests to demonstrate the utility of this design. Finally, a case study was presented to demonstrate its capacity to adapt to specific objects geometry, anticipating its potential application in situations where compliance is the priority.  
\end{abstract}
\section{INTRODUCTION}
\begin{figure}[tbp]
    \centering\includegraphics[width=1\linewidth]{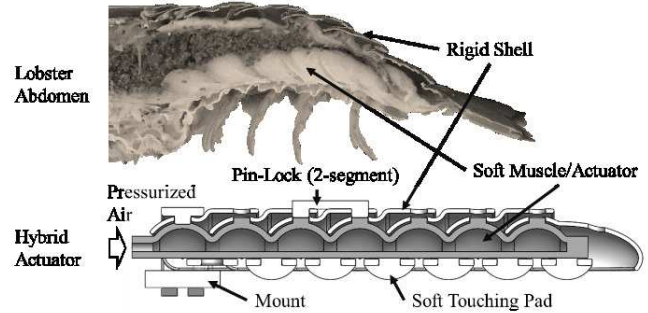}
    \caption{The illustrative design of a hybrid actuator with rigid and soft components inspired by crustaceans, which in this case is a lobster's abdomen section, commonly viewed as the "tail" section in gastronomy. Its mechanical programmability can be achieved in many ways, i.e. the selective activation of different actuator segments to generate controllable motion with different bending radius.}
    \label{fig:BioInspiration}
\end{figure}

Actuator technology has been an enabling factor in the form design of robotic systems \cite{DongjunShin2008ADesign}, which is recently moving from classical focus in robustness, precision and efficiency towards collaborative, physically safe and wearable applications \cite{Shepherd2014SoftVideo}. Recent development in soft robotics \cite{Verl2015SoftApplication} addresses this emerging need for physical human-robot interaction (pHRI) in robot designs \cite{Rus2015DesignRobots}, and special attention has been paid to bio-inspired designs to generate robotic motion and force \cite{Yi2016AAssistance}. In this paper, we draw inspirations from crustaceans, like lobsters, crabs, etc., and present a robotic actuator design through the hybridization of rigid and soft components, where the non-linear deformation of the soft actuator inside is geometrically constrained by the rigid shell outside to produce controllable yet mechanically programmable bending motions, which can be further applied to the design of a robotic finger with passive adaptation as in Fig. \ref{fig:BioInspiration}.

One major difference in robotic actuators is the mechanism of power transmission while converting energy into mechanical motion and force. Classical actuators usually employs certain rigid mechanisms to generate robotic actuation of linear and rotary motions \cite{Hollerbach1991ARobotics}. While these rigid actuators have been extensively investigated and their kinematics can be precisely predicted by well-built models, they are often made of metallic material that is heavy in weight, difficult to machine, and limited in collaborative tasks with human subjects \cite{Pervez2008SafeFuture}.

On the other hand, soft robotics has become a dynamically evolving field of research \cite{Rus2015DesignRobots} due to their inherent advantages such as compliance, light weight, low manufacturing costs and flexible designs \cite{Trimmer2013SoftRobots}. Particularly, soft pneumatic actuators (SPAs) with different structures can be designed to generate different motion profiles including bending, twisting, linear extension and contraction, and apply force or torque within the desired range \cite{Polygerinos2013TowardsRehabilitation}, making them ideal for implementation in wearable robotic devices \cite{Yap2016AImaging}. Unlike rigid actuators, soft actuators have some inherent disadvantages like low repeatability in production with the complicated fabrication process, vulnerability to ruptures and difficulty in programming the motion. In some more recent designs, solutions have been proposed to use fibers \cite{Polygerinos2015ModelingActuators} or fabric \cite{Yap2016AApplication} to constrain the radial expansion, and produce more robust prototypes with a protective layer like an extensible layer with cuts \cite{Agarwal2016StretchableDevices,Memarian2015ModellingMuscles} or origami \cite{Paez2016DesignReinforcement}. It has also been validated that varying fiber angles \cite{Connolly2015MechanicalAngle} or using sleeve coverings \cite{Galloway2013MechanicallyActuators} can modify the motion to tailor a specific task. However, it is still challenging to properly address all three issues, including repeatability, vulnerability, and  programmability, in one single SPA design.

In this study, a novel hybrid bending actuator is presented comprising both soft and rigid parts in an effort to overcome the aforementioned drawbacks and enable a more enhanced, robust, and programmable performance. Crustaceans such as lobsters use abdomens to propel through the water with a powerful tail-flip action as a startle-escape response \cite{Newland1992EscapeLobster}. The abdomen section of a lobster as shown in Fig. \ref{fig:BioInspiration} has a large lump of soft muscle enclosed by an array of rigid shells jointed in serial to produce bending motions for the tail-flip action. These jointed shells are equivalent to a serial robotic arm connected by revolute joints with parallel joint axes. The soft muscle inside takes a relatively simple shape which adapts to the inner chamber of the shell. During actuation, the soft muscle inside expands and pushes each rigid shell to move around its joint axis. Such bio-inspiration is adopted in the proposed hybrid bending actuator design with both soft and rigid components, which aims at a promising solution to address repeatability, vulnerability and programmability issues constraining the current development of soft robotics. 

In section 2, we present an in-depth description of the actuator design, its fabrication, as well as its mechanical programmability. Section 3 covers the experimental setup. Experimental results and discussion are included in section 4. In section 5, a case study is presented to demonstrate the improved compliance of hybrid actuators with reconfigurable motion. The conclusions of the study and future work are presented in the final section.

\section{Design, fabrication \&  Programmability}

\subsection{Design \& Fabrication}

Inspired by the crustaceans such as lobsters, crayfish, and mantis shrimp, the proposed hybrid actuator design consists of two parts, namely the rigid shell and the soft chamber. The soft chamber inside provides necessary actuation, while light-weight rigid shells act as a protection layer throughout the whole motion. The shell design serves multiple purposes targeted desired bending and force output in the application range. It addresses the inherent venerability issue of soft materials, and helps the soft chamber to withstand higher pressure by providing stronger constraints to the radial expansion. Especially, local buckling is avoided and a more specific bending trajectory can be predefined by the shell design, which is beneficial for analyzing and controlling. With 3D printing and a simple molding process, the hybrid actuator can be easily fabricated with minimal manual work. Moreover, mechanically programmability can be achieved by selectively activation of the joints of neighboring rigid shells which will be further discussed in the next section. 

\begin{figure}[htbp]
    \centering\includegraphics[width=1\linewidth]{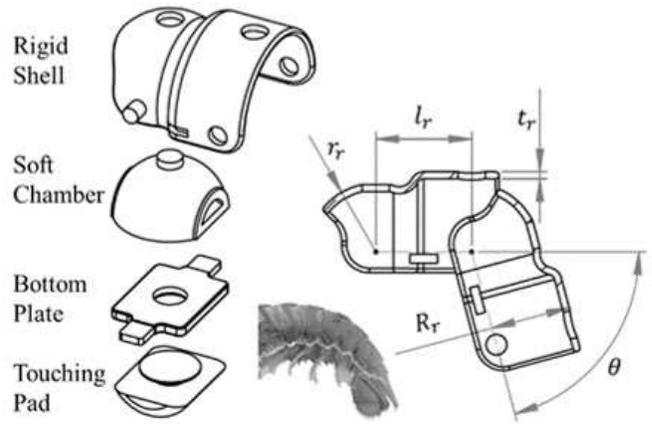}
    \caption{Rigid shell design for the body shell with five governing geometric parameters, mimicking the exoskeletons of lobster abdomen.}
    \label{fig:Design}
\end{figure}

Fig. \ref{fig:Design}  presents how the geometry of the rigid shell is governed by five parameters, including effective length ( $l_{r} =14mm$), opening angle ($\theta=75^{\circ}$), joint radius ($r_{r}=10mm$), shell radius ($R_{r}=12mm$) and wall thickness ($t_{r}=1mm$). These parameters are shared among the three types of shell designs, including the mounting shell, the body shell, and the tip shell. Fig. \ref{fig:Design} is an illustration of the body shell. The mount and tip shells are slightly modified on the outside geometry for easier fixture and force measurement. We can further extend the design to be more lobster like by decreasing the size of each shell unit towards the tip. These rigid shells were fabricated using the Form 2 SLA 3D printer by Formlabs using the standard clear resin (GPCL02). 

\begin{figure}[htbp]
    \centering\includegraphics[width=1\linewidth]{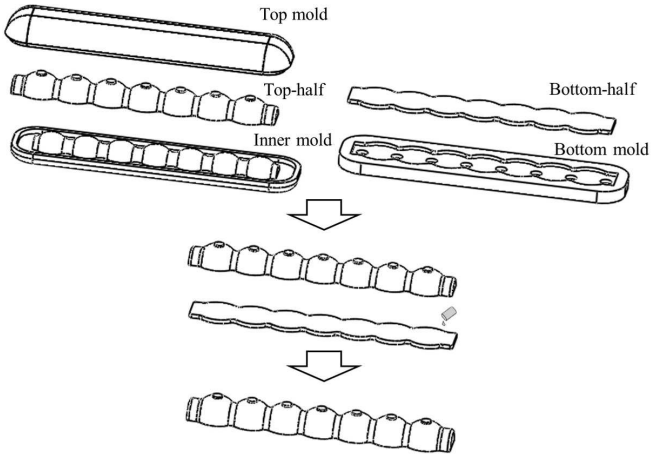}
    \caption{Two-step molding to fabricate the soft chamber for the hybrid actuator. Step 1: molding the top half of the soft chamber with the top and inner molds, while the bottom half with the bottom mold. Step 2: gluing the top and bottom halves with the same material.}
    \label{fig:Molding}
\end{figure}

Similar to the lobster muscles, the soft chamber inside the rigid shell was designed to conform to the inner walls of the rigid shell. The soft chamber was essentially simplified as a series of semi-spheres and its geometry is mainly governed by the following parameters, including the sphere radius ($r_{s}=9mm$) and chamber thickness ($t_{s}=1.5mm$). However, it should be noted that it is not necessary for the soft chamber to take the forms of the rigid shell to be functional. The soft chamber was fabricated through a simple two-step molding process, and the molds were printed with the same printer and resin as the rigid shell. As shown in Fig. \ref{fig:Molding}, three molds were 3D printed including the top mold, inner mold, and bottom mold. The top-half of the soft chamber was molded by the top and inner molds, while the bottom-half was molded by the bottom mold. A thin layer of the same mixture was added to the edge of the bottom half, and covering the top half onto the bottom half would complete the fabrication of the soft chamber. After the molding process, we can cut the length of the soft chamber to the preferred length, and then seal the two ends by placing it into a small volume of uncured silicone respectively. Once cured, a 1/8" barbed double-way connector was fed through one end of the hybrid actuator and became the mechanical connection for the pneumatic tubes. 

\begin{figure}[htbp]
    \centering\includegraphics[width=1\linewidth]{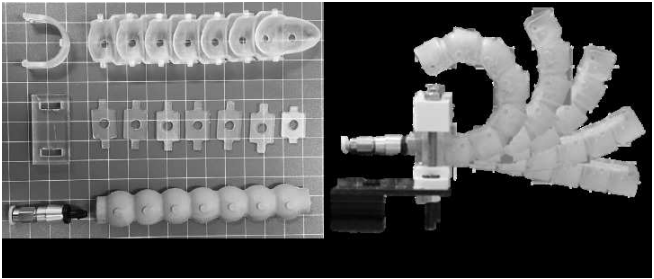}
    \caption{Assembly of the hybrid actuator with bending capability: (a) components of the hybrid bending actuator before assembly, and (b) its bending motion sequence under pressurized air.}
    \label{fig:Components}
\end{figure}

As shown in Fig. \ref{fig:Components}(a), the final assembly of the hybrid actuator involves firstly connecting a number of rigid shells in serial, then placing the soft chamber inside the rigid shells, and finally attaching the bottom positioning plates to secure the soft chamber. When pressurized air was applied to this hybrid bending actuator, it produced the motions as shown in Fig. \ref{fig:Components}(b).

\subsection{Mechanically Programmability}

\begin{figure}[htbp]
    \centering\includegraphics[width=1\linewidth]{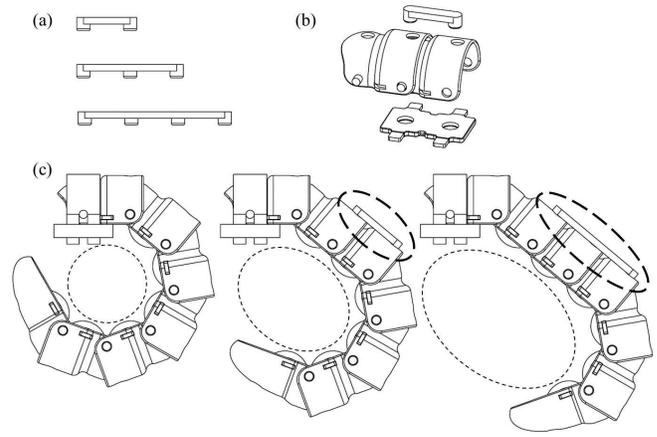}
    \caption{Mechanically programmability of the hybrid actuator's bending motion. (a) locking pins with different lengths. (b) Insertion of the locking pins on top to deactivate the rotary motion of adjacent rigid shells. (c) The reconfigured bending spaces of the hybrid actuator with 2-segment and 3-segment pin-locks inserted.}
    \label{fig:Program}
\end{figure}

The bending motion of the hybrid actuator can be reconfigured through the selective activation of the movable rigid shells. Since each rigid shell is designed as a module, we introduced a locking hole on the top of the rigid shell normal to each joint axis. As shown in Fig. \ref{fig:Program}(a), a series of locking pins with different lengths were fabricated so that they can be inserted into the holes on top of the rigid shells as in Fig. \ref{fig:Program}(b). We can also combine multiple positioning plates at the bottom to reinforce the reprogrammed actuator. As shown in Fig. \ref{fig:Program}(c), mechanically programmable bending is therefore introduced to the hybrid actuator, and desirable bending radius can be achieved that can be further applied to grasp objects and conform to different sizes and shapes. Similar mechanical programmability can be also introduced at the initial fabrication stage by changing the effective length ($l_{r}$) of the rigid shell. In this way, we can directly assemble a hybrid actuator with different bending radius, if the task is clearly defined.

\section{Experimental Setup}

\begin{figure}[htbp]
    \centering\includegraphics[width=1\linewidth]{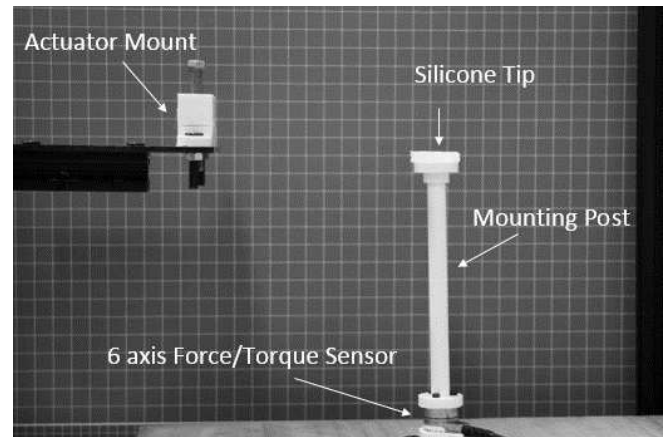}
    \caption{Experimental setup where the hybrid actuator is to be fixed on the mount with its free end pressing a post attached on top of a 6-axis force/torque sensor. }
    \label{fig:Platform}
\end{figure}

An evaluation platform (see Fig. \ref{fig:Platform}) was developed to mechanically characterize the hybrid actuators. A six-axis force/torque sensor (Nano17, ATI Industrial Automation) was used to measure the blocked force at the tip of the pressurized actuator from its neutral configuration. A recording camera (M3, Canon) was mounted on a tripod with a telephoto lens (EF 18-55mm, Canon) that was fixed at 55mm to minimize lens distortion and increase measurement accuracy. A cutting mat with checkerboard and ruler printouts was mounted directly behind and parallel to the actuator's sagittal plane, where the ruler on the mat provided a correlation between image pixels and actual length units. The open-source fluidic control board by Harvard was built to measure and control the pneumatic system, including pressure sensors, solenoid values (SMC), and miniature pump (Parker).

One hybrid bending actuator was fabricated with 7 modules measured at 105mm in total length using Elastosil M4601 by Wacker Chemical rated at 28A durometer. This silicone rubber is commonly used in soft robotics fabrication for its extensibility and high tear strength properties. It has a tensile strength of 6500kPa, 700\% elongation at break, and more than 30N/mm tear strength. Note that the first module of the actuators will be fixed to the mount. Therefore, only the rest six modules were movable with rotary motions, making it equivalent to a 6R robotic arm with all parallel joint axes, as in the first model in Fig. \ref{fig:Program}(c). 

Three pairs of pin-locks were 3D printed to mechanically program the bending motion of the hybrid actuators into 1R, 2R, and 4R configurations. The 1R configuration was achieved through a pair of pin-locks where one deactivates the rotation of three modules from the mounting end, while the other deactivated the rotation of the rest four modules from the free end. This essentially constrained the hybrid actuator with only one active revolute joint between the 3rd and 4th modules, making it a 1R manipulator with one degree-of-freedom (DoF). Similarly, the 2R and 4R configurations were achieved by using the pin-lock pair where each one deactivates the rotation of three or two modules from both the mount and free ends.

The first module of the hybrid actuator that inserted with the pneumatic connector was fixed on the evaluation platform, making the hybrid actuator cantilevered for experiments. For blocked-force measurement, the hybrid actuator was kept at zero bending angles by using a constraining platform placed on top of the hybrid actuator. Applied pneumatic pressure and the resultant force was recorded, and the actuator completed three pneumatic pressurization cycles to confirm accuracy and repeatability. For the bending test, all four configurations (1R, 2R, 4R and 6R) were tested to demonstrate the bending programmability. The camera captured the motion sequence for post-processing and analyzing the deflection angle of the distal tip segment between unpressurized and pressured state. 

\section{Results and Discussion}
The experiment results established the feasibility of the proposed hybrid actuator design that converts non-linear deformation into approximately linear bending, and revealing trade-offs in actuator performance between pneumatic pressurization and blocked forces as well as angular deflection. 

\begin{figure}[htbp]
    \centering\includegraphics[width=1\linewidth]{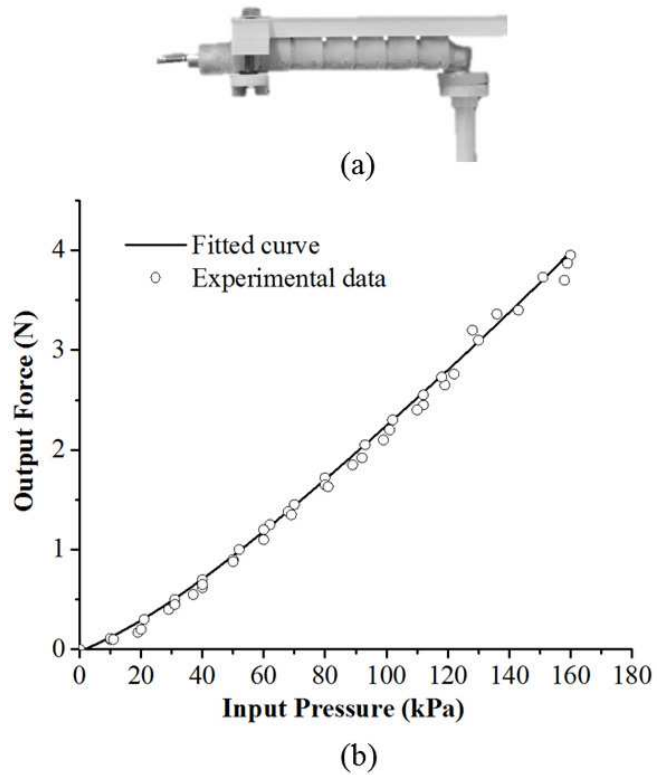}
    \caption{Blocked force test and results: (a) test setup of a 6R hybrid actuator in the force test; (b) force test result for 6R configuration.}
    \label{fig:Force}
\end{figure}

In the blocked-force test, the tip force of 6R configuration was tested (see Fig. \ref{fig:Force}(a)). The pressure was increased by an increment of approximately 10kPa, and experimental data and fitted curve for the relationship between input pressure and output force were presented in Fig. \ref{fig:Force}(b). The tip force exerted increased with increasing input pressure. An approximately linear relationship was observed, which could be caused by stronger constraints to the radial expansion provided by the shell segments. The force generated reached 4N at 165kPa. Further experimentation is undergoing to provide more throughout characterization of the hybrid actuator design and exploration into the sensitivity of each geometric parameters.

\begin{figure}[htbp]
    \centering\includegraphics[width=1\linewidth]{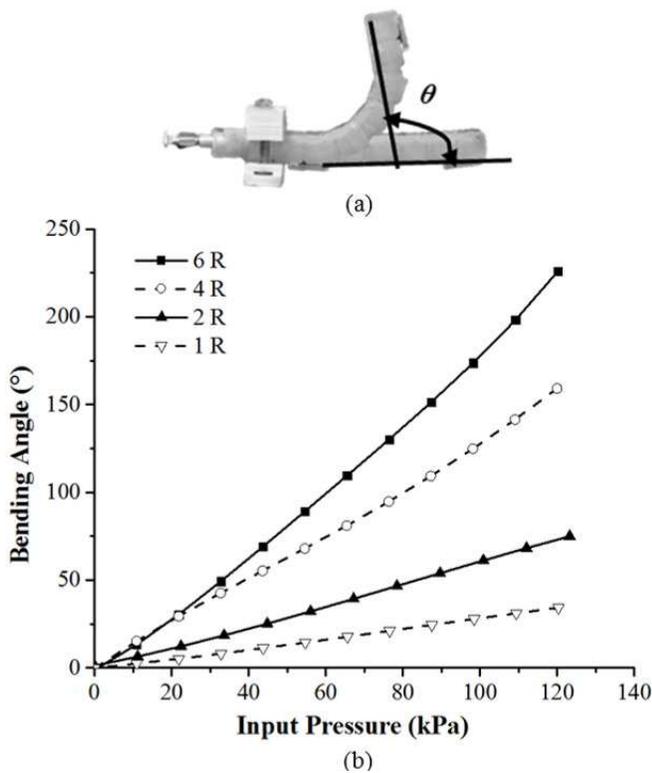}
    \caption{Bending test in different configurations and results: (a) bending angle of a 4R hybrid actuator during pressuring; (b) bending test results of the hybrid actuator of 6R, 4R, 2R and 1R configuration.}
    \label{fig:Bending}
\end{figure}

In the bending test, bending angle was defined as the deflection angle of the distal tip segment in unpressurized and pressured state (see Fig. \ref{fig:Bending}(a)). Fitted curves of relationships between bending angles and input pressure of four configurations were presented in Fig. \ref{fig:Bending}(b), and clearly configurations with higher DOFs have a stronger bending capacity. For all configurations, bending angle increased with increasing input pressure and a good linearity was also presented. In another word, the non-linear deformations generated by the soft chamber under pressurized air was constrained, yet harnessed, by the rigid shells to produce certain linearity during bending. For the 6R configuration, it reached a bending angle of $230^{\circ}$  under 120kPa, which is adequate for most wearable robotic devices. Another pilot test was also conducted using Ecoflex 00-30 by Smooth-On to fabricate the soft chamber. However, due to the ultra-softness of the silicone rubber, the hybrid actuator became so sensitive to pressure that maximum bending was reached even at 40kPa. A further test to the hybrid actuator could be conducted by choosing a different material, such as DragonSkin 10 by Smooth-on with Shore 10A softness, to horizontally compare the experimental results. 

\section{CASE STUDY}
The following case study evaluates the ability of hybrid actuators to shape match different objects. For our proposed hybrid actuator, it has adequate and reconfigurable motion. This is fundamentally different from soft actuators with infinite DOFs or rigid actuators with very limited DOFs. A can in cylindrical shape and a box in cubical shape were taken as adaptiveness tasks for they are the most common shapes in our daily life. To investigate how DOFs influence the adaptiveness of hybrid actuators, three configurations with different DOFs, including 6R, 4R, and 2R were tested to adapt to different objects under the same pressure.

\begin{figure}[htbp]
    \centering\includegraphics[width=1\linewidth]{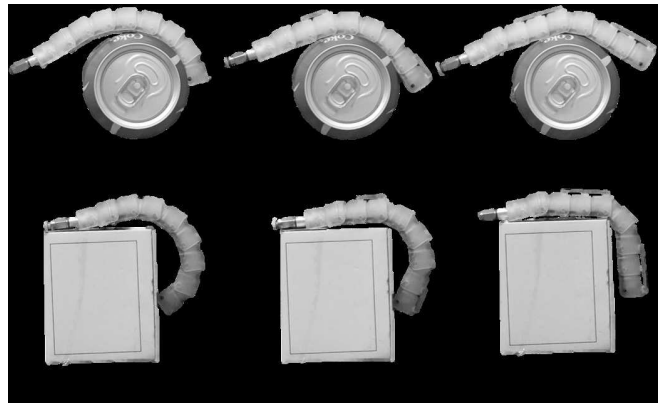}
    \caption{Adaptation tasks for hybrid actuators of 2R, 4R and 6R configurations including (a) a can in cylindrical shape and (b) a box in cubical shape.} 
    \label{fig:CaseStudy}
\end{figure}

The angular object was a cuboid box ($73mm \times 73 mm \times 87 mm$) and the circular object was a coke can (radius of 35mm), and they were fixed on the testing platform. During adaptiveness tests, the proximal tip segment of each configuration was fixed and the proximal tip segment was free to bend, and pressure was gradually increased till 100kPa inside the soft chamber. Fig. \ref{fig:CaseStudy}(a) presents the results of adaptiveness to a circular object. The 6R configuration formed into a bending curvature can be fitted with the outer circumference of the circular object, generating the best compliance. The ability of each configuration to adapt to the angular object was presented in Fig. \ref{fig:CaseStudy}(b). The 6R configuration, with most DOFs, obviously bowed at the sides of the angular object and could not properly conform to the angular corner. The 4R configuration demonstrated improved compliance as it bowed much less at the sides. The 2R configuration presented the best compliance. It suggests that more DOFs does not guarantee a better adaptiveness, and for a specific task, a proper DOFs is needed to achieve the best compliance. This is essential in wearable robotic devices or gripper for fragile objects where compliance becomes the priority in design. In the proposed hybrid design, DOFs can be easily reconfigured to improved adaptiveness to certain objects, and achieve a better compliance tailed for specific tasks, making them ideal for implementation into those applications.

\section{CONCLUSIONS}
This paper focuses on the conceptual development of a hybrid actuator aiming at some of the most urgent issues limiting the performance of soft actuators like repeatability, vulnerability and programmability. This hybrid actuator composing modular rigid shells to constrain the inflation and guide bending trajectory, and a soft chamber inside to provide the necessary actuation, which is inspired by the actuation mechanism of the abdomen of lobsters. Similar bio-inspirations can be also found in crayfish, mantis shrimp, and many other crustaceans. Unlike humans and many vertebrates that evolve with an endoskeleton design for motion generation, crustaceans utilize their exoskeletons for structure support, natural protection, as well as motion actuation. The proposed hybrid actuator can be produced in a simple fabrication process with minimal manual work. An approximately linear relationship was presented in both force and bending tests, and this might be caused by a stronger constraint of radial expansion provided by the rigid shells. The rigid shells also acted as an additional protection layer which provided full protection for the soft chamber even under high pressures. A simple method to rapidly program the bending curvature was proposed by using pins to selectively inactivate some shells instead of modifying the mechanical structures \cite{Galloway2013MechanicallyActuators}. This leads to adjustable DOFs of hybrid actuators which could present better compliance tailored for specific tasks. 

Different from the natural geometry of the lobster abdomen, the hybrid actuator presented in this paper is intentionally designed to be longer and thinner, making it closer to the shape factor of human fingers. Combined with the mechanical programmability introduced in this paper, one potential integration of these hybrid actuators can be found in the design and development of robotic glove for hand rehabilitation. The hybrid actuator adopted can generate precise and intricate motions regarding custom requirements. Future work will continue to provide further design review and investigation to improve the force output of the hybrid actuator, and sensitivity research especially in the selection of suitable rigid and soft materials and parameters for fabrication. 
 
\addtolength{\textheight}{-2cm}   

\bibliographystyle{IEEEtran}
\bibliography{root}
\end{document}